\documentclass[letterpaper, 10 pt, conference]{ieeeconf}  

\IEEEoverridecommandlockouts                              

\usepackage{cite}
\usepackage{amsmath,amssymb,amsfonts}
\usepackage{algorithmic}
\usepackage{graphicx}
\usepackage{textcomp}
\usepackage{epsfig} 
\usepackage{mathtools}
\usepackage{color}
\usepackage{etoolbox}
\usepackage{xpatch}
\usepackage{multirow}
\usepackage{booktabs}
\let\labelindent\relax
\usepackage{enumitem}
\usepackage{comment}
\usepackage{caption}
\usepackage{subcaption}
\usepackage{eucal}
\usepackage{threeparttable}

\newcommand{\cmt}[1]{}

\long\def\ignorethis#1{}

\newcommand{\figref}[1]{Fig.~\ref{#1}}
\renewcommand{\eqref}[1]{(\ref{#1})}

\newcommand{\vc}[1]{\ensuremath{\pmb{#1}}}

\newcommand{\set}[1]{\ensuremath{\mathcal{#1}}}

\newcommand{\pctab}{\hspace{0.2in}}

\usepackage{amsthm}
\theoremstyle{definition}

\title{\bf Hybrid Learning- and Model-Based Planning and Control\\of In-Hand Manipulation}
\author{Rana Soltani Zarrin$^{1}$, Katsu Yamane$^{2}$, Rianna Jitosho$^{3}$
\thanks{$^{1}$Rana Soltani Zarrin is with Honda Research Institute USA, San Jose, CA.
        {\tt\footnotesize rana\_soltanizarrin@honda-ri.com}}%
\thanks{$^{2}$Katsu Yamane is with Path Robotics Inc, Columbus, OH. This work was done while at Honda Research Institute USA.
        {\tt\footnotesize katsu.yamane@gmail.com}}%
\thanks{$^{3}$Rianna Jitosho is with the Mechanical Engineering department of Stanford University, Stanford, CA. Contribution to this work was done during an internship at Honda Research Institute USA.
        {\tt\footnotesize rjitosho@stanford.edu}}%
}

\makeatletter
\xpatchcmd{\@thm}{.}{}{}{}
\makeatother

\begin{document}
\maketitle
\thispagestyle{empty}
\pagestyle{empty}

\begin{abstract}
This paper presents a hierarchical framework for planning and control of in-hand manipulation of a rigid object involving grasp changes using fully-actuated multifingered robotic hands.
While the framework can be applied to the general dexterous manipulation, we focus on a more complex definition of in-hand manipulation, where at the goal pose the hand has to reach a grasp suitable for using the object as a tool.
The high level planner determines the object trajectory as well as the grasp changes, i.e. adding, removing, or sliding fingers, to be executed by the low-level controller. While the grasp sequence is planned online by a learning-based policy to adapt to variations, the trajectory planner and the low-level controller for object tracking and contact force control are exclusively model-based to robustly realize the plan. By infusing the knowledge about the physics of the problem and the low-level controller into the grasp planner, it learns to successfully generate grasps similar to those generated by model-based optimization approaches, obviating the high computation cost of online running of such methods to account for variations. 
By performing experiments in physics simulation for realistic tool use scenarios, we show the success of our method on different tool-use tasks and dexterous hand models. Additionally, we show that this hybrid method offers more robustness to trajectory and task variations compared to a model-based method.
\end{abstract}

\section{Introduction}

\begin{figure}
     \centering
     \begin{subfigure}[b]{0.15\textwidth}
         \centering
         \includegraphics[width=\textwidth]{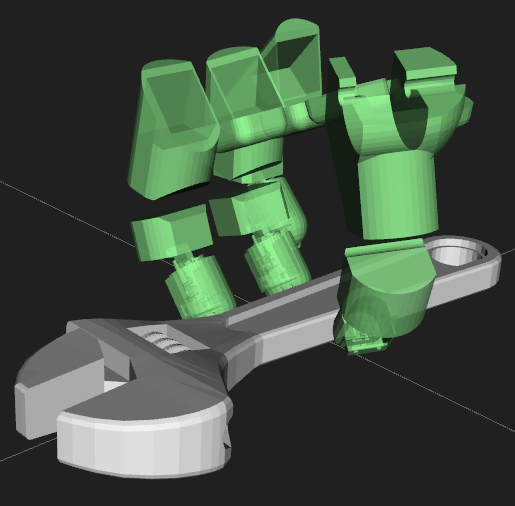}
     \end{subfigure}
     \vspace{0.1cm}
     \begin{subfigure}[b]{0.15\textwidth}
         \centering
         \includegraphics[width=\textwidth]{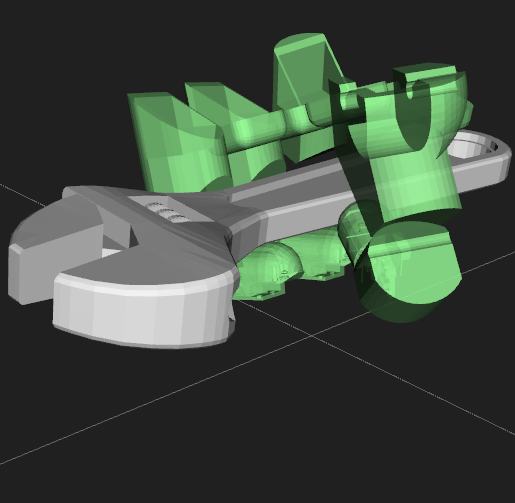}
     \end{subfigure}
     \caption{Example of tool manipulation. Left: initial grasp for picking up; right: final grasp for using the tool.}
     \label{fig:in-hand-example}
\end{figure}

To use a tool for its designed purpose, it often has to be held with a grasp that is different from the one for picking it up.
To turn a nut using a wrench, for example, one would first pick up the wrench using the fingertips and then pull it closer to the palm while transitioning to power grasp so that a large force can be applied (\figref{fig:in-hand-example}). 
It is therefore necessary to change the grasp along with the object pose relative to the hand between picking up and using the tool.

While similar to the in-hand manipulation problem, which focuses on object {\em reposing}, the tool use problem poses a constraint on the grasps that can be used. While for the object reposing any grasp that can realize the goal pose can be used, the final grasp in the tool-use problem should enable applying a specific wrench (force/torque) to the object.

In this paper, we present a planning and control framework for in-hand manipulation that can also enable the more complex dexterous manipulation problem of tool-use.

\section{Related Work}

In order for the finger joints to stay within their limits while reposing an object in-hand, finger gaiting, sliding, or rolling a link along the object surface could be needed.
While some of the past works utilized external forces such as the gravitational force and contact forces from the environment, e.g. supporting the object by the hand palm or reposing the object through non-prehensile primitives~\cite{dafle2014extrinsic,chavan2015prehensile,karayiannidis2015hand,van2015learning,nonprehensile_liu}, we focus on in-hand manipulation using only internal forces.
We also consider generic fully-actuated robotic hands without assuming any underactuation or mechanical compliance that can help improve the stability and robustness~\cite{liarokapis2017deriving,abondance2020dexterous, van2015learning}.

Earlier works employed model-based approaches such as constrained optimization~\cite{liu2009dextrous,mordatch_CIO,nonprehensile_liu} to determine the hand and/or object trajectories as well as finger-object contact interactions for simple object reorientation tasks. 
While these methods can generate plausible in-hand manipulation motions including sliding and rolling for a pre-determined task, they are computationally expensive to be run online and do not consider real-time feedback control and therefore are not suitable for robotics applications.

Since dynamics of the hand-object system may be difficult to model or have large uncertainties, a data-driven learning-based approach is a promising alternative.
This approach can further be divided into learning-from-demonstration and reinforcement learning (RL).
An example of the former is Ueda et al.~\cite{ueda2010multifingered} where a cylinder reposing was achieved using direct teaching from human demonstrations.
However, in addition to dependency on costly demonstration data, grasp changes would be more challenging to teach due to the differences between human and robotic hands.
Model-free RL methods can learn a policy without extensive modeling effort or human demonstrations. Taking advantage of extra stability resulted by supporting the object through palm, multiple end-to-end Deep RL (DRL) methods have been used for in-hand reorientation of simple objects such as a cylinder or cube \cite{kumar2016optimalControlRL,andrychowicz2020OpenAI_dexterousManipulation}.
Attempts have been made to improve the sample efficiency of these methods by augmenting demonstrations~\cite{rajeswaran2017DRL-demonstration} or using model-based RL~\cite{nagabandi2020model-based-RL}.

Hybrid learning- and model-based method is a possible approach for realizing both sample efficiency and robustness.
Li et al. \cite{Hierarchical_Control} proposed a hierarchical control structure for the typical in-hand manipulation where a learned policy determines the motion primitives (sliding, flipping, reposing) and a model-based low-level controller executes the selected primitive. They evaluated the method by a simulated 3-fingered hand with 2 degrees of freedom (DoF) on each finger, reposing a pole and cube in a 2D vertical plane. 

This paper presents a hierarchical hybrid learning- and model-based dexterous manipulation planning and control framework for applications that require not only object reposing but also a grasp suitable for tool-use, which has been rarely considered in prior work.
An RL policy infused with domain knowledge of the physics of the problem and the low-level controller outputs a sequence of finger joint addition, removal, or sliding actions in real-time, that are robustly realized by the model-based controller. 
We speculate that a hybrid structure enables more data-efficient learning~\cite{Hierarchical_Control} compared to end-to-end RL approaches~\cite{andrychowicz2020OpenAI_dexterousManipulation,kumar2016optimalControlRL}, although we have not conducted a formal comparison on the amount of data required for training.  We also show that, unlike a model-based approach, this hybrid approach can enable adapting to variations.
We demonstrate our method in a simulated physics environment on realistic hands with 16~DoF (Honda dexterous hand \cite{hasegawa2022MFH} and Allegro hand\cite{Allegro}) and realistic tool models. Hardware evaluation is underway.

\section{Problem Definition}
\label{sec:overview}

\subsection{Assumptions}

This paper concerns the problem of in-hand manipulation of a single rigid object using a dexterous robotic hand with at least four fingers in order to change the grasp while maintaining 3D force closure.
We assume that inertial and kinematic properties of the object and hand, as well as the friction coefficient between the hand links and object, are known.
We also assume that every joint is torque controlled in both directions, and that each link is equipped with a tactile or force-torque sensor that gives the total 3-dimensional force and center of pressure of the distributed force applied to the link surface, determining the contact point. Currently, we also assume the object manipulation happens through the fingers only (i.e. no palm contact).
Finally, we assume that the object's pose (position and orientation) is given by a technique such as vision-based object tracking.

\subsection{Planning Phase}
\label{planning overview}
To define the planning problem, we first define a tuple called contact information $C = \{J, \vc{c}_J, \vc{c}_O\}$ where $J$ denotes the joint (link) in contact with the object, and $\vc{c}_J$ and $\vc{c}_O$ are the contact point location represented in the joint and object frames respectively. We assume a point contact on each link.
A grasp $G = \{C_1, C_2, \ldots\}$ is defined as a set of 0 or more contact information.
If $G = \varnothing$, there is no contact and therefore the object is not held.
To facilitate planning, we provide a set of possible grasp candidates $\set{G}_{cand}$.

Given these definitions, the in-hand manipulation planning problem is defined as follows:
\theoremstyle{definition}
\newtheorem*{definition*}{Definition}
\begin{definition*}[In-hand manipulation planning]
  Given
  \begin{itemize}
  \item a set of grasp candidates $\set{G}_{cand}$
  \item initial grasp $G_s\in \set{G}_{cand}$,
  \item initial and final object positions $\vc{p}_s$ and $\vc{p}_g$,
  \item initial and final object orientations $\vc{R}_s$ and $\vc{R}_g$,
  \item external wrench $\vc{w}_{ext}$ applied to the object at $(\vc{p}_g, \vc{R}_g)$
  \end{itemize}
  find
  \begin{itemize}
  \item execution time $T$,
  \item object reference position $\hat{\vc{p}}_O(t)$ and orientation $\hat{\vc{R}}_O(t)$ as a function of time $t\in [0, T]$, and
  \item sequence of grasps $G_m\;(m=1,2,\ldots,M)$ where $M>1$ is a user-defined integer representing the number of uniformly distributed sampling times  $t_m =$\footnotesize$Tm/(M-1)$\normalsize
  \end{itemize}
  such that
  \begin{itemize}
  \item the object does not collide with the stationary part of the hand (i.e. palm) or the environment, 
  \item the contact points in $G_m$ are reachable, and
  \item the contacts in $G_m$ are able to provide the wrench required to generate the acceleration at $t=t_m$ along the object reference trajectory $\hat{\vc{p}}_O(t),\hat{\vc{R}}_O(t)$.
  \end{itemize}
\end{definition*}

The object trajectory is computed in two steps:
\begin{enumerate}
\item Path planning: obtain a collision-free path of the object such that every waypoint has at least one grasp in $\set{G}_{cand}$ in which all contact points can be reached.
\item Trajectory generation: obtain $\hat{\vc{p}}_O(t), \hat{\vc{R}}_O(t)$ by optimizing the timestamp of each waypoint obtained in step~1). The timestamp of the last waypoint becomes the completion time $T$. Also compute $\hat{\vc{p}}_{Om} = \hat{\vc{p}}_O(t_m)$ and $\hat{\vc{R}}_{Om} = \hat{\vc{R}}_O(t_m)\;(m=1,2,\ldots,M)$.
\end{enumerate}

\begin{figure}
\begin{center}
\includegraphics[scale=0.35]{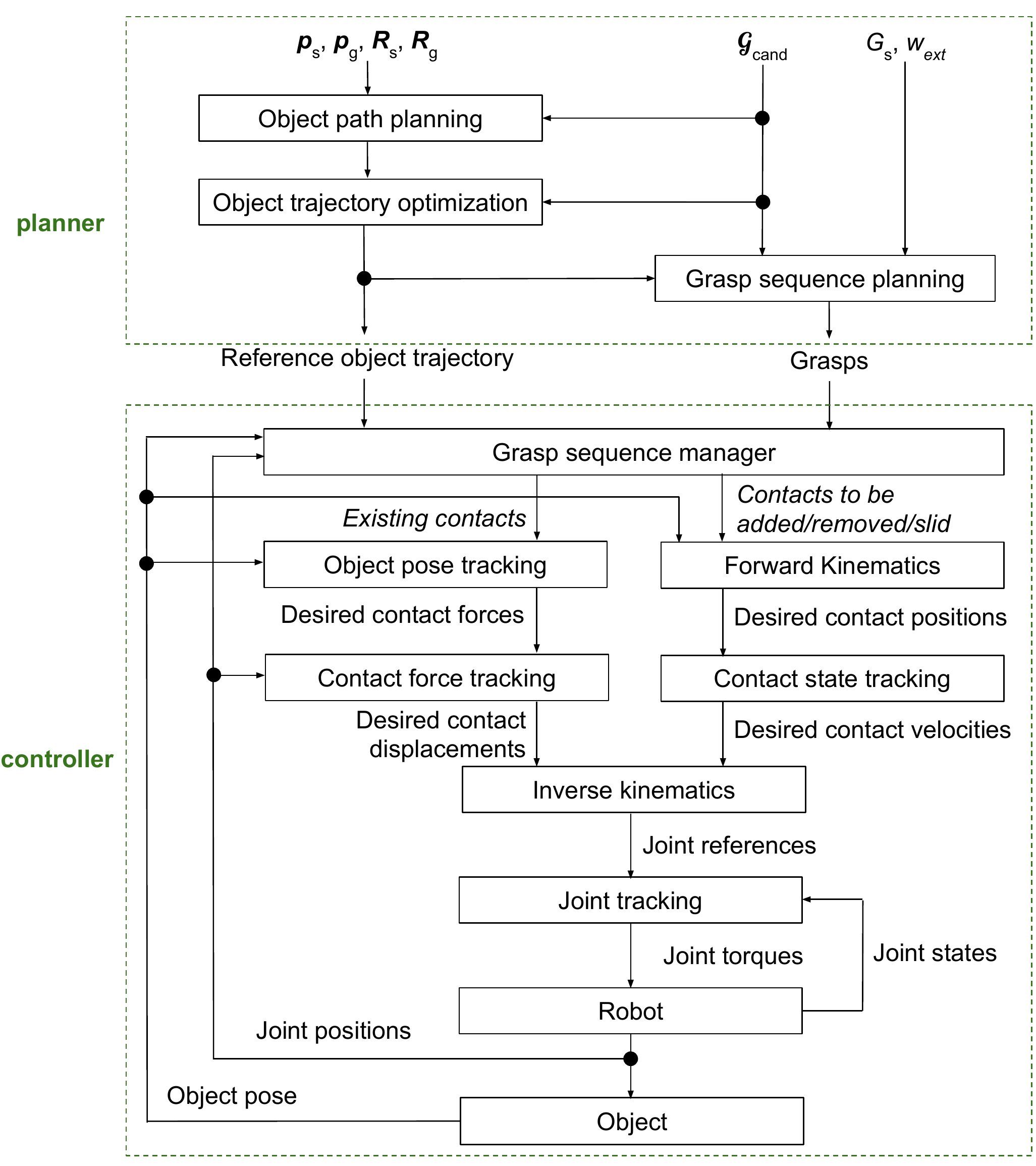}
\end{center}
\caption{In-hand manipulation framework}
\label{fig:in-hand-manipulation-framework}
\end{figure}

Fig.~\ref{fig:in-hand-manipulation-framework} includes the structure of the planning phase. Details of the planners are provided in section \ref{sec:planning}.

\subsection{Control Phase}

Structure of the controller, which executes the in-hand manipulation plan generated in the planning phase, is shown in Fig. \ref{fig:in-hand-manipulation-framework}. The {\em grasp sequence manager} block runs with a constant time interval and determines whether a grasp transition should take place.
For instance with a learning-based method, the policy is called to determine if a contact should be added, removed, slid, or the current grasp should be maintained.

The remainder of the controller consists of following three low-level tracking controllers (see Section~\ref{sec:controller} for details):
\begin{enumerate}
\item Object tracking: compute the contact forces applied to the object such that it tracks the planned trajectory,
\item Contact force tracking: compute the joint reference positions to realize the contact forces from 1), and
\item Contact state tracking: realize the contact states dictated by the planned contact sequence.
\end{enumerate}

\section{Preliminaries}
\label{sec:preliminaries}


\subsection{Inverse Kinematics}
\label{sec:ik}

To compute the joint positions $\vc{q}$ for a given grasp with $K$ contacts, 
$C_k = \{J_k,\vc{c}_{Jk},\vc{c}_{Ok}\}\; (k=1,2,\ldots,K)$, and object pose $(\vc{p}_O, \vc{R}_O)$,
we apply an iterative algorithm:
\begin{gather}
\Delta \vc{q} = \arg\min \sum_{k=1}^K ||\vc{J}_{Ck}\Delta \vc{q} - \Delta \vc{p}_k ||^2\label{eq:qp-ik}
\end{gather}
subject to element-wise bounds
\begin{equation}
\frac{1}{k_{IK}}(\vc{q}_{min} - \vc{q}) \leq \Delta\vc{q} \leq \frac{1}{k_{IK}}(\vc{q}_{max} - \vc{q})
\end{equation}
where
\begin{eqnarray}
\Delta \vc{p}_k &=& (\vc{p}_O + \vc{R}_O \vc{c}_{Ok}) - \left(\vc{p}_{Jk}(\vc{q}) + \vc{R}_{Jk}(\vc{q}) \vc{c}_{Jk}\right)
\label{eq:ik-position-error}
\end{eqnarray}
and $k_{IK}$ is a positive gain, $\vc{J}_{Ck}$ is the Jacobian matrix of contact point $k$ with respect to $\vc{q}$, $\vc{p}_{Jk}(\vc{q})$ and $\vc{R}_{Jk}(\vc{q})$ denote the pose of joint $J_k$ at $\vc{q}$, and $\vc{q}_{max}$ and $\vc{q}_{min}$ are the vector of maximum and minimum joint positions respectively.

Using the total IK error defined as $d(\vc{q}, \vc{p}_{O}, \vc{R}_O, G) = \sum_{k=1}^K ||\Delta\vc{p}_k ||^2$, iteration terminates with $\vc{q}^*$ and $d^*$ if:
\begin{itemize}
    \item $d(*)$ is below a predefined threshold, or
    \item $d(*)$ increases from the previous iteration, or
    \item any of the links make a contact with the environment.
\end{itemize}
The maximum IK error for grasp $G$ at sample $m$ is:
\begin{equation}
\Delta p^*(m, G) = \max_{k=1,2,\ldots,K} ||\Delta \vc{p}_k ||^2.
\end{equation}

\subsection{Contact Force Optimization}
\label{sec:contact-force-optimization}

Using the Newton-Euler equations of 3D rigid-body dynamics, we can compute the total force and torque to be applied to the object moving with angular velocity $\vc{\omega}_O$ to generate linear and angular accelerations $(\dot{\vc{v}}_O,\; \dot{\vc{\omega}}_O)$ by
\begin{eqnarray}
\hat{\vc{f}}_{total} &=& {\mathcal M} \dot{\vc{v}}_{O} - \vc{f}_E\\
\hat{\vc{\tau}}_{total} &=& {\mathcal I}\dot{\vc{\omega}}_{O} + \vc{\omega}_{O} \times {\mathcal I} \vc{\omega}_{O} - \vc{\tau}_E
\end{eqnarray}
where ${\mathcal M}$ and ${\mathcal I}$ are the object mass and moments of inertia and $\vc{f}_E$ and $\vc{\tau}_E$ are the applied external force and torque.

Given a grasp $G$ we optimize the contact forces $\vc{f}_k$:
\begin{equation}
\vc{f}^*_1,\vc{f}^*_2,\ldots,\vc{f}^*_K = \arg\min Z_f \label{eq:qp-force}
\end{equation}
subject to constraints
\begin{eqnarray}
\boldsymbol{c}^T_{kl}\boldsymbol{f}_k \leq 0 \; (k=1,2,\ldots,K;~ l=1,2,\ldots,L) \label{eq:friction_inequality_const}
\end{eqnarray}
where
\begin{equation}
\resizebox{0.85\columnwidth}{!}{%
$Z_f = ||\hat{\vc{f}}_{total} - \sum_{k=1}^K \vc{f}_k ||^2 + w_t ||\hat{\vc{\tau}}_{total} - \sum_{k=1}^K \vc{p}_{Ok} \times \vc{f}_k||^2$}
\end{equation}
$\vc{p}_{Ok} = \vc{R}_{Om} \vc{c}_{Ok}$, $L$ is the number of sides of the pyramid approximating the friction cone, $w_t>0$ is a user-defined weight, and $\vc{c}_{kl}$ is the the normal vector of the $l$-th side of the pyramid at contact $k$ which can be computed as
$\vc{c}_{kl} = \left(
\vc{t}_{1k}\; \vc{t}_{2k}\; \vc{n}_k
\right) \left(
\cos \theta_l\; \sin \theta_l\; -\mu \right)^T$
where $\mu$ is the friction coefficient, $\vc{t}_{1k}$ is a tangent vector at contact $k$, $\vc{t}_{2k} = \vc{t}_{1k}\times \vc{n}_k$, and $\theta_l = 2\pi l / L$. The inequality constrain (\ref{eq:friction_inequality_const}) applies to the sticking contacts. For a sliding contact, we constrain the $f_k$ to be on the edge of the friction cone, with the tangent component in the direction of desired sliding.
We also define the following variables related to contact forces:
\begin{equation}
\hat{f}_{total}(m) = || \hat{\vc{f}}_{total}||^2 + w_t|| \hat{\vc{\tau}}_{total}||^2
\end{equation}
\begin{equation}
\resizebox{0.95\columnwidth}{!}{%
$e^*(m, G) = ||\hat{\vc{f}}_{total} - \sum_{k=1}^K \vc{f}^*_k ||^2+ w_t ||\hat{\vc{\tau}}_{total} - \sum_{k=1}^K \vc{p}_{Ok} \times \vc{f}^*_k||^2$}
\end{equation}
\begin{equation}
f^*(m, G) = \sum_{k=1}^K ||\vc{f}^*_k ||^2
\end{equation}
where the argument $m$ indicates that the object pose, velocity and acceleration at sample $m$ are used.
Essentially, $e^*(m,G)$ represents the residual wrench that grasp $G$ cannot generate.
We use the ALGLIB library~\cite{bib-alglib} to solve the quadratic programs of \eqref{eq:qp-ik} and \eqref{eq:qp-force}.

\section{In-Hand Manipulation Planning}
\label{sec:planning}

The initial grasp can be found by a grasp planner~\cite{bohg2013data} or prior knowledge from demonstrations.
In our implementation, we generate $\set{G}_{cand}$ by choosing 1--3 possible contact points for each middle and distal joint, and then enumerating all combinations of contacts that do not cause collisions.

\subsection{Object Path Planning}
We chose a sampling-based motion planning algorithm based on Probabilistic Roadmap (PRM)~\cite{kavraki1996prm} called PRM*~\cite{karaman2011prmstar} since it worked best for our problem.
PRM builds a roadmap of valid samples in the configuration space by random sampling.
A valid path can be found by searching for a path that connects the start and goal configurations in the roadmap.
PRM* extends PRM to find an optimal path.
Our implementation uses the Open Motion Planning Library~\cite{sucan2012the-open-motion-planning-library}.

The configuration space is a 6D space representing the 3D position $\vc{p}$ and orientation $\vc{R}$ of the object.
A sampled configuration is valid if the object pose satisfies the following:
\begin{enumerate}
\item The object does not collide with the environment (floor) or the fixed part (palm) of the robot hand.
\item There exists at least one grasp $G$ in which $d^*(\vc{p}, \vc{R}, G)$ is smaller than a threshold.
\end{enumerate}

In this work, we use the path length as the cost function to be minimized.

\subsection{Object Trajectory Generation}
\label{sec:trajectory-optimization}

Let $N$ denote the number of waypoints of the path obtained by path planning.
The goal is to determine the timestamp $t_i$ of waypoint $i\;(i=1,2,\ldots,N)$ with constraints $t_1=0$, $t_{i-1} + \Delta t_{min} \leq t_i \;(2 \leq i \leq N)$ and $t_N \leq T_{max}$, where $\Delta t_{min} > 0$ is the minimum interval between waypoints and $T_{max}$ is the maximum duration.

Given a set of timestamps, we can interpolate the waypoints with piecewise cubic B-splines such that the trajectory passes the initial and final poses with zero velocity.
As a result, we obtain 7 sets of cubic B-splines for the 3 position components and 4 quaternion components.
We then use these B-splines to sample the whole trajectory at $M$ sample points with a uniform time interval $t_N / (M-1)$.

We formulate the problem of determining the timestamps as a numerical optimization problem with cost function
\begin{equation}
Z_1 = \sum_{m=1}^M c^*(m)
\end{equation}
where $c^*(m)$ is the cost at sample $m\;(1\leq m \leq M)$, obtained as $c^*(m) = \min_{G\in \set{G}_{cand}} c(m,G)$ where $c(m,G)$ is the cost for using grasp $G$ at sample $m$. The interpolated trajectory gives the position $\vc{p}_{Om} = \vc{p}_O(t_m)$, orientation $\vc{R}_{Om}=\vc{R}_O(t_m)$, linear and angular velocities and accelerations $\vc{v}_{Om}$, $\vc{\omega}_{Om}$, $\dot{\vc{v}}_{Om}$ and $\dot{\vc{\omega}}_{Om}$ at sample $m$. Using these quantities, $c(m,G)$ is computed by
\begin{equation}
\resizebox{0.88\columnwidth}{!}{%
$c(m, G) = d^*(\vc{p}_{Om},\vc{R}_{Om},G) + w_e e^*(m,G) + w_f f^*(m,G) \label{eq:sample-set-cost}$}
\end{equation}
where $w_e$ and $w_f$ are user-defined weights.

In our implementation, we use the Constrained Optimization by Linear Approximations (COBYLA) algorithm implemented in the NLopt library~\cite{bib-nlopt}.

\subsection{Online Grasp Sequence Planning}
\label{sec:learning}

We use DRL to find a policy that can determine the grasp which can enable realizing the desired object trajectory while counteracting the external wrench $\vc{w}_{ext}$.
In the current implementation the grasp generated by the planner differs from the previous grasp only one join at a time.
Also, while $\vc{w}_{ext}$ could be time dependant in general to represent external disturbances experienced by the tool, in this work we limit $\vc{w}_{ext}$ to the final expected wrench to be counteracted by tool.
We formulate the Grasp Sequence planning problem as a Markov Decision Process (MDP) dependent on the output of the trajectory optimization problem as follows: 

\begin{itemize}[leftmargin=*]
    \item {\bf State $\vc{s}_m$: } Robot joint positions $\vc{q}$, reference tool pose $(\hat{\vc{p}}_O(m), \hat{\vc{R}}_O(m))$, goal pose $(\hat{\vc{p}}_g,\hat{\vc{R}}_g)$, current sample $m$, current grasp $G_m$, external expected wrench $\vc{w}_{ext}$.
    \item {\bf Action $\vc{a}_m$: } grasp change command selected from the following discrete action space:
\begin{align}
\scriptsize
   &\mathcal{A} = \{command (C)\ | \thinspace C \in \mathcal{G}\ and \\
   &command \in \{(add, remove, slide \thinspace to, no \thinspace change)\} \nonumber
\end{align}
\normalsize
which consists of $\sum_{k=1}^{N_J} (2n_{C(k)}+1)+1$ actions, where $n_{C(k)}$ is the number of contact candidates on link $k$. 
    \item {\bf Reward: } Some $R<<0$ if the tool or robot comes into collision with the environment (\ref{reward_weights}). Otherwise, calculated based on the metrics determining kinematic realizability of the new grasp, ability of the commanded grasp $\hat{G}(m)$ to realize the desired object trajectory $(\hat{\vc{p}}_O(m), \hat{\vc{R}}_O(m))$ and to counteract the external wrench $\vc{w}_{ext}$:
\begin{multline} \label{eq:reward}
    R(\vc{s}_m, \vc{a}_m) = -w_1 \Delta p^*(m,\hat{G}_m)- w_2 e^*(m,\hat{G}_m)\\ + \frac{w_3 \hat{f}_{total}(m)}{f^*(m,\hat{G}_m)}
    - t(m,G_m)-s(m,G_m,\hat{G}_{m})
\end{multline} 
where $w_i > 0$ ($i=1,2,3)$ are user-defined weights.  
The first two terms encourage the agent to avoid actions that result in large IK or wrench error, and the third term rewards grasps that require smaller contact forces. The domain knowledge about the physics of the problem and low-level controller is infused to the learning side through the above defined reward terms. $t(m,G_{m})$ is a penalty to discourage invalid transitions:
\begin{equation}
\resizebox{0.9\columnwidth}{!}{%
$  t(m,G_{m}) = \left\{
  \begin{array}{ll}
    2 & \mbox{if } a_m=add(C) \; || \; slide(C) \; \wedge \; C \in G_{m}\\
    2 & \mbox{if } a_m=remove(C) \; \wedge \; C \notin G_{m}\\
    10 & \mbox{if } a_m=slide(C) \; \wedge \; C(J) \notin G_{m}\\
    0 & \mbox{otherwise.}
  \end{array}
  \right. $}
\end{equation}
The last term is a sliding specific penalty, to encourage gaiting over sliding when the sliding distance is large or when contacts are on different surfaces of the object:
\begin{equation}
    s(m,G_m,\hat{G}_{m}) = w_4 \Delta s+w_5 \theta_n
\end{equation}
where $\Delta s$ is the sliding distance and $\theta_n$ is the angle between  object normals at the two contact points.

\item {\bf Transition: } The system dynamics proceeds to $s_{m+1}$ according to reference trajectory and updates the grasp according to action $a_m$. The episode ends if there are less than two links in contact with the object, or if the maximum timestep is reached.
\end{itemize}

\section{Controller}
\label{sec:controller}

\subsection{Object Tracking} \label{Object_tracking}

The desired object acceleration given the current ($\vc{p}_{O}$, $\vc{R}_{O}$) and reference object pose ($\hat{\vc{p}}_O(m)$, $\hat{\vc{R}}_O(m)$) is:
\begin{eqnarray}
\hat{\dot{\vc{v}}}_{O} &=& k_{P1} (\hat{\vc{p}}_O(m) - \vc{p}_{O}) - k_{D1} \hat{\vc{v}}_{O}\\
\hat{\dot{\vc{\omega}}}_{O} &=& k_{P2}\Delta \vc{r}_{O} - k_{D2} \hat{\vc{\omega}}_{O}
\end{eqnarray}
where $k_{P1}$, $k_{D1}$, $k_{P1}, k_{D2} > 0$ are feedback gains, $\Delta \vc{r}_{O}$ is a vector given by $\vc{a}\sin \theta$ where $\vc{a}$ and $\theta$ are the rotation axis and angle to transform $\vc{R}_{O}$ to $\hat{\vc{R}}_O(m)$, and $\hat{\vc{v}}_{O}$ and $\hat{\vc{\omega}}_{O}$ are obtained by integrating $\hat{\dot{\vc{v}}}_{O}$ and $\hat{\dot{\vc{\omega}}}_{O}$ respectively.
We do not use measured object velocity because it is likely to be noisy.
Instead, we use the integration of desired object acceleration for the damping term.
Using $\hat{\dot{\vc{v}}}_O$, $\hat{\dot{\vc{\omega}}}_O$ and $\hat{\vc{\omega}}_O$, we optimize the contact forces for the given grasp using (\ref{eq:qp-force}).

\subsection{Contact Force Tracking}
\label{sec:contact-force-tracking}

The optimized contact forces $\vc{f}^*_k$ are tracked by a controller similar to admittance control as follows.
Let us assume that contact point $k$ is on the $j$-th finger and there is no other contact point on the finger, and let $\vc{J}_{Ckj}$ denote the Jacobian matrix of the position of contact point $k$ with respect to the joint positions of finger $j$. 
The contact force error at a contact point $k$ can be compensated by joint torque:
\begin{equation}
\Delta \vc{\tau}_j = \vc{J}^T_{Ckj} (\vc{f}^*_k - \vc{f}_k)
\end{equation}
which can be produced by reference joint offset:
\begin{equation}
\Delta \vc{q}_j = \vc{K}^{-1}_{Pj} \Delta \vc{\tau}_j.
\label{eq:joint-position-displacement}
\end{equation}
where $\vc{K}_{Pj}$ denotes a diagonal matrix whose elements are the proportional gains of the joints of finger $j$.

Directly adding $\Delta \vc{q}_j$ to the current joint reference positions will cause an issue if we want the object to track a given trajectory because the finger also has to follow the object motion.
To solve this issue, we add the contact point displacement due to the object motion as well:  
\begin{equation}
\Delta \vc{p}_k = \vc{J}_{Ckj} \Delta \vc{q}_j + \Delta t (\hat{\vc{v}}_{O} + \hat{\vc{\omega}}_{O} \times \vc{p}_{Ok})
\end{equation}
which can be used in place of \eqref{eq:ik-position-error} by a single iteration of the IK algorithm in Section~\ref{sec:ik} to obtain the new joint reference position, which will be tracked by a proportional-derivative controller with gravity compensation.

\subsection{Contact State Change}

A grasp change involves adding a new contact, moving a contact point while maintaining the contact, e.g. sliding, or removing an existing contact.

To add a new contact, we use the IK algorithm in Section~\ref{sec:ik} to move the desired contact point on the finger toward the contact point on the object. 
Once a contact force is detected, the finger is controlled to maintain a small constant contact force (0.1~N in our experiments) using the force tracking controller described in Section~\ref{sec:contact-force-tracking} until the contact force is maintained above a threshold (0.05~N) for a given duration (0.1~s).
The reference object pose is fixed until the new contact is established.

For sliding towards a new contact point, the IK algorithm in (\ref{eq:ik-position-error}) is modified by projecting $\Delta \vc{p}_k$ onto the object surface. 
\begin{equation}
\Delta \vc{p}_k := \Delta \vc{p}_k - (\vc{n}_k^T \Delta \vc{p}_k) \vc{n}_k
\end{equation}
where $\vc{n}_k$ is the normal vector at $C_k$.
A hybrid force and position controller realizes the sliding by performing force control in the normal direction to maintain contact while performing position control on the sliding direction.

When an existing contact needs to be removed, the finger is controlled by the IK algorithm to move the contact point away from the object in the normal vector direction.
The new grasp is declared to be established when no force at the contact to be removed is detected for a given period (1.0~s).

\section{Experiments}
\label{sec:experiments}

\subsection{Experimental Setup}

\begin{figure}
     \centering
     \begin{subfigure}[b]{0.11\textwidth}
         \centering
         \includegraphics[width=\textwidth]{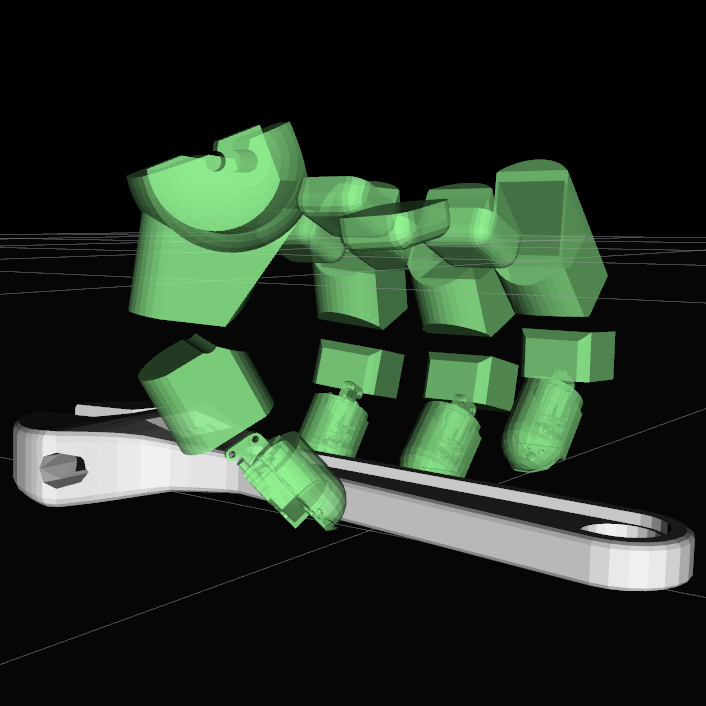}
     \end{subfigure}
     \vspace{0.1cm}
     \begin{subfigure}[b]{0.11\textwidth}
         \centering
         \includegraphics[width=\textwidth]{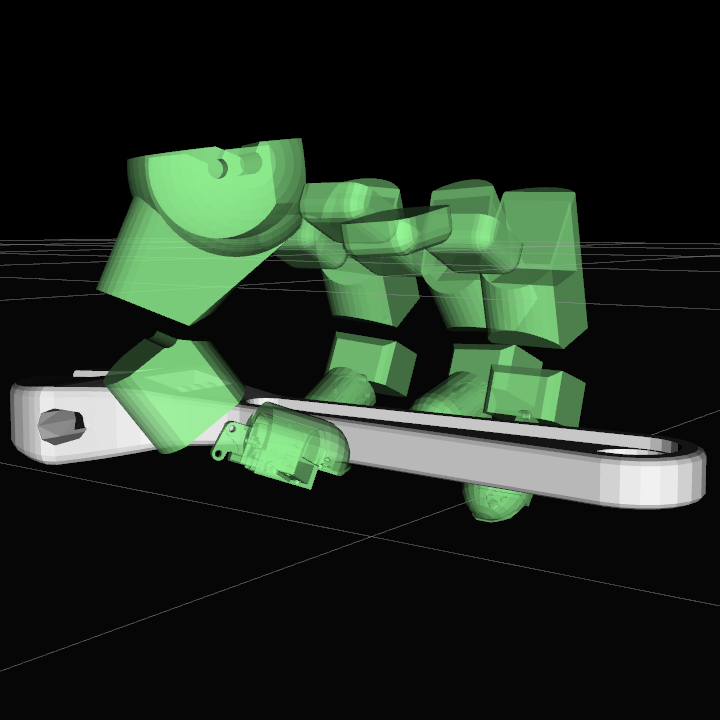}
     \end{subfigure}
    \begin{subfigure}[b]{0.11\textwidth}
         \centering
         \includegraphics[width=\textwidth]{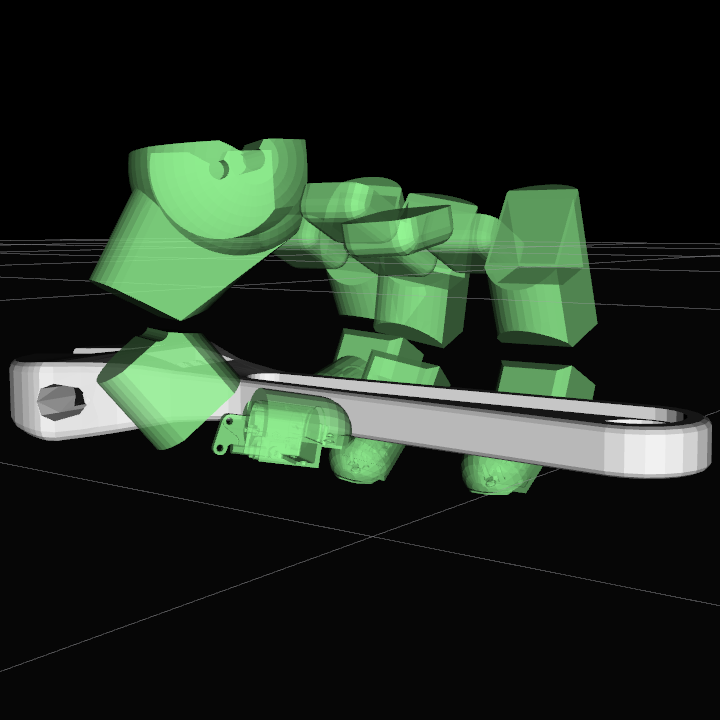}
     \end{subfigure}
     \begin{subfigure}[b]{0.11\textwidth}
         \centering
         \includegraphics[width=\textwidth]{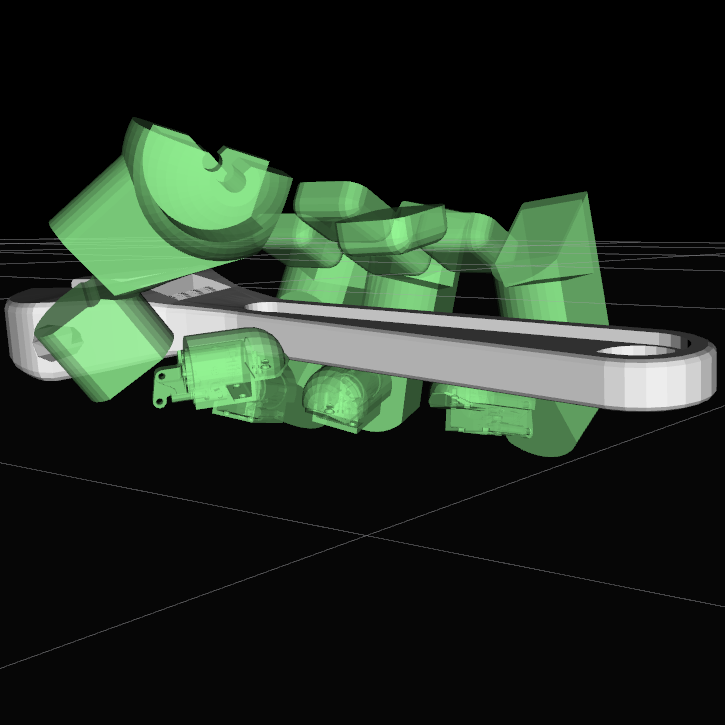}
     \end{subfigure}

     \begin{subfigure}[b]{0.11\textwidth}
         \centering
         \includegraphics[width=\textwidth]{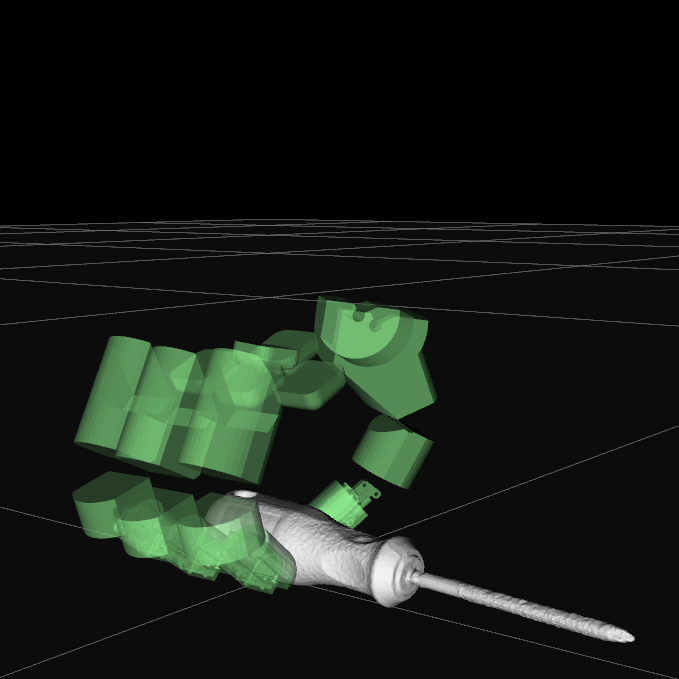}
     \end{subfigure}
     \begin{subfigure}[b]{0.11\textwidth}
         \centering
         \includegraphics[width=\textwidth]{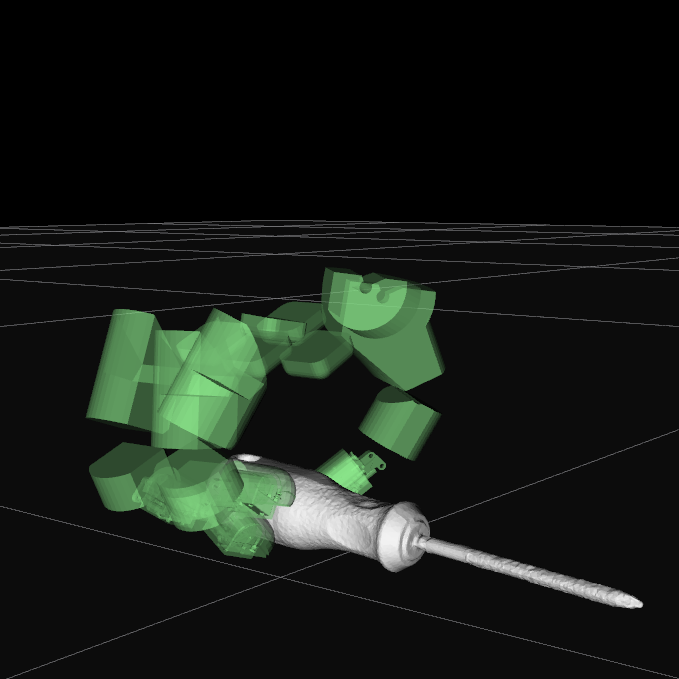}
     \end{subfigure}
    \begin{subfigure}[b]{0.11\textwidth}
         \centering
         \includegraphics[width=\textwidth]{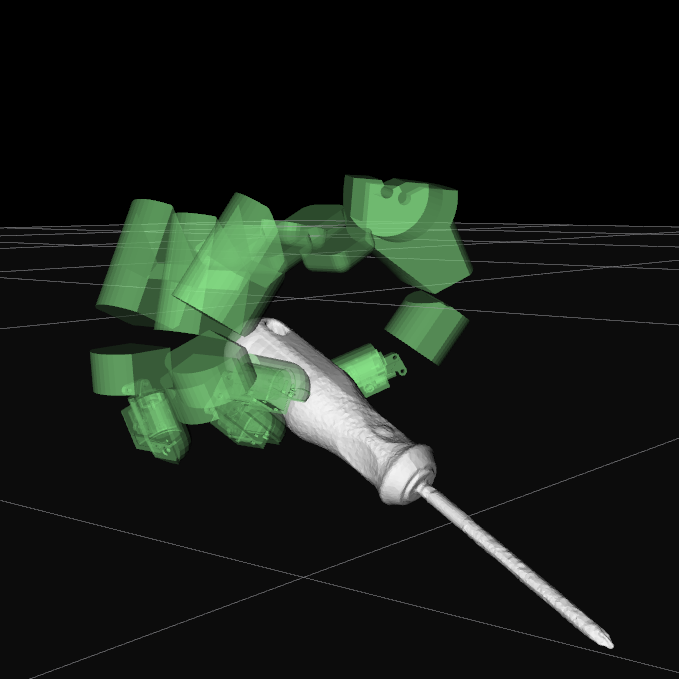}
     \end{subfigure}
     \begin{subfigure}[b]{0.11\textwidth}
         \centering
         \includegraphics[width=\textwidth]{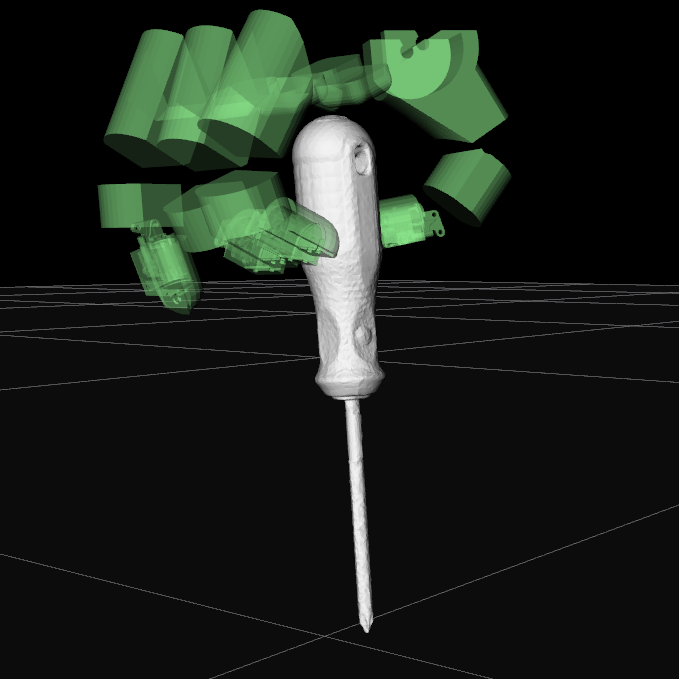}
     \end{subfigure}
        \caption{Sample sequence of grasp changes resulted by the framework, from left (initial grasp) to right (final grasp)}
        \label{fig:grasp_sequence}
\end{figure}

We validate the planning and control framework in simulation using two representative in-hand manipulation tasks with realistic tool models.
In both tasks, the objects are initially lying horizontally on a table (\figref{fig:grasp_sequence}). The nominal tasks are:
\begin{itemize}
    \item Wrench: pick up with prismatic 2-finger grasp~\cite{feix2015grasp_taxonomy} and lift about 0.08~m while gradually transitioning to power grasp that wraps the index and middle fingers around the wrench to apply a large torque to turn a nut.
    \item Screwdriver: pick up with prismatic 3-finger grasp and lift about 0.13~m and rotate $90^\circ$ to take a vertical pose while transitioning to tripod grasp so that a screw placed vertically on the table can be rotated.
    In this example, the hand is also lifted about 0.14~m.
\end{itemize}

We use an anthropomorphic robotic hand model with 4 fingers, each consisting of 4~DoF corresponding to the 2~DoF of human MCP joint and 1~DoF each of the PIP and DIP joints\cite{hasegawa2022MFH}.
The simulator computes the contact forces using the contact model proposed by Todorov\cite{todorov2011convex} with the friction cone of friction coefficient $\mu=1$ approximated by a $L=12$-sided pyramid, and then computes the joint accelerations using unit vector method~\cite{walker1982efficient}, which are integrated by 4-th order Runge-Kutta integrator at 1~ms timestep.
The low-level controller (Section~\ref{sec:controller}) and grasp sequence manager (\figref{fig:in-hand-manipulation-framework}) run at 100~Hz and 10~Hz respectively on separate threads.

We train the learning-based planner using Proximal Policy Optimization (PPO) \cite{schulman2017PPO} with 15 parallel environments using the implementation in TF-Agents~\cite{TFAgents}. Discount factor is selected as 0.99. For the entropy regularization coefficient and learning rate hyperparameters, we select higher values at the beginning of the training (0.5 and $5\times10^{-4}$ respectively) to allow more exploration and faster learning~\cite{ahmed2019entropy}. 
We decrease both parameters linearly as the training progresses to ensure convergence to the optimal solution.
To improve robustness, we perform domain randomization~\cite{tobin2017domain} during training by adding random variations in the range of [-0.01, 0.01]~(m) for position and [-0.1, 0.1]~(rad) for orientation to the given initial $(\vc{p}_s,\vc{R}_s)$ and final $(\vc{p}_g,\vc{R}_g)$ object poses. For the wrench example, we train two policies with different external torques around the vertical axis applied at the last sample (i.e. loosening and tightening tasks): policy $RL_{-1}$ with $-1$\thinspace Nm and $RL_{1}$ with $1$\thinspace Nm, while we train one screwdriver policy without external torque, since the final desired grasp for this task does not depend on the direction of rotation of the screwdriver.
Each policy is trained in $10^6$ iterations with 15 episodes each (equivalent to 347 days of experience), which takes about 4 days on a desktop computer with NVIDIA\textsuperscript{\textregistered} Quadro\textsuperscript{\textregistered} P2200 GPU. The real-time running of the policy takes about 1~sec.
Equation (\ref{reward_weights}) shows the weights used for the reward terms in (\ref{eq:reward}).

As the baseline, we compare our method with a model-based approach that plans the optimal sequence of grasps offline using dynamic programming (DP) with the same cost function as the RL. DP takes about 150\!~sec to compute on the same machine. Such a long computation time, makes such approaches unsuitable for real-time implementation.

\scriptsize
\begin{align}\label{reward_weights}
  w_1 &= \left\{
  \begin{array}{ll}
    100/\Delta p^*(m,G_m) & \mbox{if } \Delta p^*(m,G_m)\geq 10^{-4}\\
    200 & \mbox{otherwise} 
  \end{array}
  \right.\\
  w_2 &= \left\{
  \begin{array}{ll}
    100/e^*(m,G_m) & \mbox{if } e^*(m,G_m) \geq 1\\
    50 & \mbox{otherwise} \nonumber
  \end{array}
  \right.\\
  w_3 &= \left\{
  \begin{array}{ll}
    2000 & \mbox{if } e^*(m,G_m) < 1 \wedge (||\vc{\tau}_{Em}||>0 \vee ||\vc{f}_{Em}|| >0)\\
    10 & \mbox{if } e^*(m,G_m) < 1 \wedge (||\vc{\tau}_{Em}||=0 \wedge ||\vc{f}_{Em}|| =0)\\
    0 & \mbox{otherwise} \nonumber
  \end{array}
  \right.\\
  w_4 &= 10 \nonumber \\ 
  w_5 &= \left\{
  \begin{array}{ll}
  \theta_n/2 & \theta_n<100^\circ\\ 
  10 & \mbox{otherwise} \nonumber
  \end{array}\right.
\end{align}
\normalsize
\subsection{Qualitative Evaluation}
\begin{figure}
    \centering
    \includegraphics[scale=0.09]{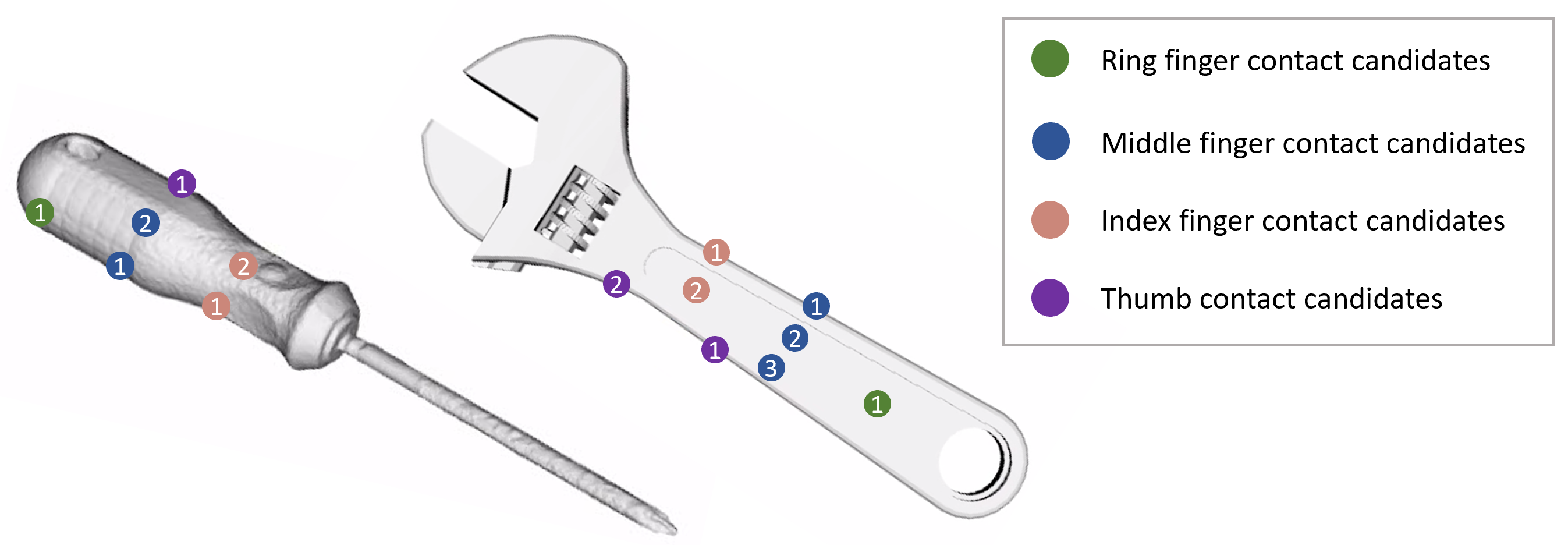}
    \caption{Object contact points}
    \label{fig:contacts}
\end{figure}

When evaluated in the simulation, the proposed approach successfully performed the desired tool-use tasks (videos attached). A comparison of the generated grasp sequences for some example tasks is shown in Table~\ref{Tab:grasp sequence}, where n:no\thinspace change, a:add, r:remove, s:slide, T:thumb, I:index, M:middle, R:ring, d:distal\thinspace link, m:middle\thinspace link, and subscripts show the contact candidates. The $-1$ and $1$ subscripts in the policy denote the applied external torque, and g means gaiting\thinspace only (i.e. no sliding primitive). See Fig. \ref{fig:contacts} for the current choice of contact locations on the tools (wrench is reversed to show the points on the bottom surface). As can be seen, for identical tasks, RL generates sequences similar to that of DP, which shows that RL is capable of generating an optimal policy. In the wrench task, for example, both methods maintain the initial grasp until the sixth sample when the wrench is elevated enough for the fingers to be placed beneath the wrench without colliding with the floor. When sliding primitive is enabled, RL carefully decides between gaiting and sliding ($RL_{-1}$). Adding sliding primitive speeds up the learning process $30-50\%$ by reducing the actions to learn.

We also observe that RL planner can result in different final grasps for different external torques, verifying the ability of the planner to choose a suitable grasp for the task. For example, when the external torque applied to the wrench simulates the requirement for loosening the nut, thumb is placed further away from the wrench head, whereas for tightening the nut the thumb sits closer to the wrench head.

By applying the algorithm to Allegro hand we showed that the method is transferable to different hand models.

\begin{table}
\centering
\caption{Commanded grasp changes for different tasks starting from their initial grasps.}
\label{Tab:grasp sequence}
\resizebox{\linewidth}{!}{
\begin{tabular}{@{\hskip0pt}l@{\hskip2pt}|l@{\hskip2pt}|@{\hskip2pt}l@{\hskip0.5pt}}
\hline
task & policy & grasp sequence\\
\hline\hline
\multirow{4}{*}{\textrm{wrench}} & $DP_{g_{-1}}$ &   [$n$, $n$, $n$, $n$, $n$, $a(R_{d_1})$, $r(M_{d_1})$, $a(M_{d_2})$, $n$, $r(I_{d_1})$, $a(I_{d_2})$, $a(I_{m_1})$, $r(M_{d_2})$, $a(M_{d_3})$, $a(M_{m_1})$, $n$]\\
\cline{2-3}
& $RL_{g_{-1}}$ &  [$n$, $n$, $n$, $n$, $n$, $a(R_{d_1})$, $r(M_{d_1})$, $a(M_{d_2})$, $r(I_{d_1})$, $a(I_{d_2})$, $n$, $n$, $a(I_{m_1})$, $r(M_{d_2})$, $a(M_{d_3})$, $a(M_{m_1})$] \\
\cline{2-3}
& $RL_{-1}$ &  [$n$, $n$, $n$, $n$, $n$, $a(R_{d_1})$, $s(M_{d_2})$, $n$, $r(I_{d_1})$, $a(I_{d_2})$, $n$, $n$, $a(I_{m_1})$, $s(M_{d_3})$, $a(M_{m_1})$, $n$] \\
\cline{2-3}
& $RL_{1}$ &  [$n$, $n$, $n$, $n$, $n$, $a(R_{d_1})$, $s(M_{d_2})$, $n$, $r(I_{d_1})$, $a(I_{d_2})$, $a(I_{m_1})$, $s(M_{d_3})$, $a(M_{m_1})$, $n$, $s(T_{d_2})$, $n$]\\
\hline
\textrm{screw}& $DP$ &  [$n$, $s(M_{d_2})$, $n$, $n$, $n$, $n$, $n$, $r(R_{d_1})$, $s(I_{d_2})$, $n$, $n$, $n$, $n$, $n$, $n$]\\
\cline{2-3}
\textrm{driver} & $RL$ &  [$n$, $n$, $n$, $n$, $n$, $r(R_{d_1})$, $n$, $n$, $n$, $n$, $n$, $n$, $s(M_{d_2})$, $s(I_{d_2})$, $n$, $n$]\\
\hline
\end{tabular}
}
\end{table}

\subsection{Robustness Evaluation}

We analyze the robustness of each method by adding random variations in the ranges of [-0.03,0.03]~(m) for the horizontal position and [-20,20]~(deg) for the yaw rotation of the start pose, and [-0.025, 0.025]~(m) for the horizontal position and [-0.03, 0.01]~(m) for the vertical position of the goal pose.
We divide the trials into two groups: {\em medium variation} if all variations are within [-0.01, 0.01]~(m) for position and [-10,10]~(deg) for orientation, and {\em large variation} otherwise.
We run 50 trials for each object-method pair (25 in each variation group).
A trial is a failure if one or more of the following events occur: 1) the object is dropped, 2) the hand makes an unintended collision with the object or environment, 3) the final object pose has a position error larger than 0.005~m or an orientation error larger than 0.1~rad

\begin{table}
\centering
\caption{Robustness comparison between the proposed hybrid method and model-based method for different tasks} \label{Tab:success rate}
\resizebox{\linewidth}{!}{
\begin{tabular}{l|c|c|c|c|c}
\hline
\multirow{3}{*}{task} & \multirow{3}{*}{planner} &
\multicolumn{2}{c|}{medium variation} & \multicolumn{2}{c}{large variation}\\
\cline{3-6}
& & success & orientation error  & success & orientation error \\
& & rate & (radians)  & rate & (radians)
\\\hline
\multirow{2}{*}{wrench} & RL\textsubscript{1}   & 64\%  & 0.060 & 48\%  & 0.207 \\
& DP & 40\% & 0.085 & 20\% & 0.341\\
\hline
\multirow{2}{*}{screwdriver}& RL  & 76\%  & 0.068 & 56\%  & 0.085\\
& DP   & 84\%  & 0.103 & 40\%  & 0.257\\
\hline
\end{tabular}
}
\end{table}

Table~\ref{Tab:success rate} summarizes the success rate and the orientation error at the end of the trajectory under the applied external torque, for each 25 trial for each combination of task, planner, and variation level. Position errors are negligible due to no external force in the current experiments.

In both variation groups, RL has higher robustness thanks to realtime adaptation of the grasp sequence as well as domain randomization during training.  Naturally, the success rate decreases as variation increases.
In addition to not being exposed to variations of such magnitude during planning or training, another possible reason is that some of the desired contact points no longer could be reachable.
Resultantly, the realized grasp will have large contact position errors which in turn can cause large wrench errors and tilting or dropping the object.
Possible solutions for this include using a larger set of possible contact points or non-prehensile manipulation primitives during the planning phase, and adaptively modifying the contact points during control.

\section{Conclusion}
\label{sec:conclusion}

In this paper, we presented a framework for robust and data-efficient in-hand tool manipulation where in addition to object reposing, achieving a final grasp that enables tool-use is required. The learning-based grasp sequence planner that is infused with knowledge about the physic of the problem and the low-level controller, can successfully infer the optimal contact transitions to robustly be realized by the controller and can react to variations introduced during run-time. We conducted simulation experiments in two in-hand manipulation tasks using realistic four-fingered robotic hands and object models and showed that the approach can successfully be applied to different hands, objects, and tasks. Future work includes comparison of performance and data efficiency to end-to-end learning and hardware implementation ~\cite{Allegro,hasegawa2022MFH}.

\bibliographystyle{IEEEtran}
\bibliography{main}

\end{document}